\documentclass{article}

    \PassOptionsToPackage{numbers, compress}{natbib}

\usepackage[final]{optrl_2019} 

\usepackage[utf8]{inputenc} 
\usepackage[T1]{fontenc}    

\usepackage{url}            
\usepackage{booktabs}       
\usepackage{amsfonts}       
\usepackage{nicefrac}       
\usepackage{microtype}      

\usepackage{amsmath}
\usepackage{amssymb}
\usepackage{amsthm}
\usepackage{enumitem} 

\usepackage{algorithm}
\usepackage[noend]{algpseudocode}
\usepackage{amssymb,mathtools,amsfonts}
\algrenewcommand\algorithmicindent{1em}%
\usepackage{hyperref}       

\usepackage{verbatim}

\newcommand{\inner}[3]{\left\langle #1,#2\right\rangle_{#3}}

\newcommand{\norma}[2]{\|\,#1\,\|_{#2}}

\newcommand{\om}{\mu ^{\pi}}

\newcommand{\sign}{{\rm sign}}

\newcommand{\x}[2]{x_{#1}^{#2}}

\newenvironment{proofof}[1]{\begin{proof}[{#1}]}{\end{proof}}

\newcommand{\ie}{\emph{i.e.}\ }
\newcommand{\iid}{i.i.d.\ }

\newcommand{\mcf}{\mathcal}
\newcommand{\mbf}{\mathbf}

\DeclareMathOperator{\Exp}{\mbf{E}}
\DeclareMathOperator{\Prob}{\mbf{P}}
\newcommand{\ones}{\mbf{1}}
\renewcommand{\Re}{\mbf{R}}

\newcommand{\Nat}{\mbf{N}}

\newcommand{\sspace}{\mathcal{X}}     
\newcommand{\aspace}{\mathcal{A}}     

\newcommand{\eps}{\varepsilon}

\newcommand{\tpose}{^T}

\newcommand{\abs}[1]{|#1|}
\newcommand{\norm}[1]{\lVert#1\rVert}

\newcommand{\innerprod}[2]{\left\langle{#1},{#2}\right\rangle}

\newcommand{\Lem}{{Lem.}}

\newcommand{\Thm}{{Thm.}}

\newtheorem{lemma}{Lemma}
\newtheorem{proposition}{Proposition}

\newtheorem{theorem}{Theorem}

\newtheorem{assumption}{Assumption}

\title{Stochastic convex optimization for provably efficient apprenticeship learning
	\thanks{{This project has received funding from the European Research Council 
			(ERC) under the European Union’s Horizon 2020 research and innovation 
			programme grant agreement OCAL, No. 787845.}}}

\author{%
  Angeliki Kamoutsi \\
  ETH Zurich\\
 \texttt{kamoutsa@ethz.ch} \\
   \And
  Goran Banjac \\
  ETH Zurich\\
 \texttt{gbanjac@ethz.ch}  \\
 \And
  John Lygeros \\
  ETH Zurich\\
  \texttt{jlygeros@ethz.ch} \\
}

\begin{document}

\maketitle

\begin{abstract}
  We consider large-scale Markov decision processes (MDPs) with an unknown cost function and employ stochastic convex optimization tools to address the problem of imitation learning, which consists of learning a policy from a finite set of expert demonstrations.
  We adopt the apprenticeship learning formalism, which carries the assumption that the true cost function can be represented as a linear combination of some known features. Existing inverse reinforcement learning algorithms come with strong theoretical guarantees, but are computationally expensive because they use reinforcement learning or planning algorithms as a subroutine. On the other hand, state-of-the-art policy gradient based algorithms (like IM-REINFORCE, IM-TRPO, and GAIL), achieve significant empirical success in challenging benchmark tasks, but are not well understood in terms of theory. With an emphasis on non-asymptotic guarantees of performance, we propose a method that directly learns a policy from expert demonstrations, bypassing the intermediate step of learning the cost function, by formulating the problem as a single convex optimization problem over occupancy measures. We develop a computationally efficient algorithm and derive high confidence regret bounds on the quality of the extracted policy, utilizing results from stochastic convex optimization and recent works in approximate linear programming for solving forward MDPs.
 
\end{abstract}

\section{Introduction}\label{sec:introduction}
The goal of apprenticeship learning (AL) in a Markov decision process (MDP) environment without cost function is to learn a policy that achieves or even surpasses the performance of a policy demonstrated by an expert. A usual assumption is that the unknown true cost function can be  represented  as  a  weighted  combination  of  some  known  basis  functions, where the true unknown weights specify how different desiderata should be traded off. An argument for this assumption is that in practice the unknown cost function depends on just a few key properties, but the desirable weighting is unknown. 

A lot of methods have been proposed to solve the AL problem. The most naive approach is \emph{behavior cloning}, which casts the problem as a supervised learning problem, in which the goal is to learn a map from states to optimal actions. Although behavior cloning is simple and easy to implement, the crucial i.i.d. assumption made in supervised learning is violated. As a result, the approach suffers from the problem of \emph{cascading errors} which is related to \emph{covariate shift}~\cite{Ho:2016}.  Later works like DAgger\cite{ross2011reduction} eliminate this distribution mismatch by formulating the problem as a no-regret algorithm in an online learning setting. However, these algorithms require interaction with the expert, which is a different learning scenario from the one considered in this paper. Most importantly, their sample and computational complexities scale polynomially with the horizon of the problem, which in our case is infinite.

Inverse reinforcement learning (IRL)~\cite{Abbeel:2004} is a prevalent approach to AL. In this paradigm, the learner first infers the unknown cost function that the expert tries to minimize and then uses it to reproduce the optimal behavior. IRL algorithms do not suffer from the problem of cascading errors because the training takes place over entire expert trajectories, rather than individual actions. In addition, since the recovered cost function ``explains'' the expert behavior, they can easily generalize to unseen states or even new MDP environments. Note however, that most existing IRL algorithms \cite{Abbeel:2004,Syed:2007,Ng:2000,Abbeel:2008,Ziebart:2008,Ratliff:2006,Neu:2007,Levine:2010,Levine:2011} are computationally expensive because  they  use  reinforcement  learning  as  a  subroutine.  

On the other hand, one can frame the problem as a single convex program~\cite{syed2008apprenticeship}, bypassing the intermediate step of learning the cost function. Although the associated program can be solved exactly for small-sized MDPs, the approach suffers from the curse of dimensionality, making it intractable for large-scale problems, arising in, e.g., autonomous driving with increasing number of sensors and decision aspects. Provably efficient convex approximation schemes for the convex formulation of AL~\cite{syed2008apprenticeship} in the context of large-scale MDPs remain unexplored. However, it is worth noting, that the formulations and reasoning in~\cite{syed2008apprenticeship} formed the ground and inspired later state-of-the-art algorithms~\cite{Ho:2016,Ho:2016b}. In particular, the authors in~\cite{Ho:2016} developed a gradient-based optimization formulation over parameterized policies for AL, and then presented algorithms which are parallel to the policy gradient RL counterparts \cite{Williams:1992,Schulman:2015}. The sequel paper \cite{Ho:2016b} draws a connection between the policy optimization formulation and generative adversarial networks \cite{Goodfellow:2014}, from which an analogous imitation learning algorithm is derived. These approaches are model-free and scale to large and continuous environments. However, in general the policy optimization problem is highly non-convex and as a result remains  hampered by limited theoretical understanding. In particular, these methods provide no guarantees on the quality of the points they converge to. 

With an emphasis on non-asymptotic guarantees of stability and performance, in this work we propose an approximation scheme for the convex formulation of AL~\cite{syed2008apprenticeship}.  In particular, the objective is to minimize the $\ell_1$-distance between the feature expectation vector of the expert and the learner, subject to linear constraints ensuring that the optimization variable is an occupancy measure induced by a policy. Our AL algorithm and its theoretical analysis build upon recent innovations in approximate linear programming (LP) for large-scale discounted MDPs~\cite{Abbasi-Yadkori:2014,Yadkori2019} and can be seen as the AL analogue of their algorithhms.
Similar to \cite{Abbasi-Yadkori:2014}, we control the complexity by limiting our search to the linear subspace defined by a small number of features.
We then convert the initial program to an unconstrained convex optimization problem.
To this end, we use a surrogate loss function by adding a multiple of the total constraint violations to the initial objective.
We then construct unbiased subgradient estimators and apply the stochastic subgradient descent algorithm.
In this way, by combining bounds in the stochastic convex optimization literature and concentration inequalities, we are able to give high confidence regret bounds showing that the performance of our algorithm approaches the best achievable by any policy in the comparison class. A salient feature of the algorithm is that the iteration and sample complexities do not depend on the size of the state space but instead on the number of approximation features.

It is worth mentioning that, since our methodology is based on the LP formulation of MDPs \cite{Hernandez-Lerma:1996,Hernandez-Lerma:1999,Hernandez-Lerma:2002,Borkar:1988}, it is naturally extensible to unconventional problems involving additional safety constraints or secondary costs, where traditional dynamic programming techniques are not applicable \cite{Hernandez-Lerma:2003,Dufour:2013, Shafieepoorfard:2013}.

To the best of our knowledge this is the first time that a performance bound is derived for a policy-optimization-based algorithm for AL.  We hope that the techniques proposed in this work provide a starting point for developing provably efficient AL algorithms.

\textbf{Notation.\,\,\,}
We denote by $A_{i,:}$ and $A_{:,j}$ the $i$-th row and $j$-th column of a matrix $A$, respectively. For $p \in [1,\infty]$, we denote by $\|\cdot\|_{p}$  the $p$-norm in $\mbf{R}^n$. The corresponding induced matrix norm is defined by $\norma{A}{p}=\sup_{\norma{x}{p}\le 1}\norma{Ax}{p}$. For vectors $x$ and $y$, we denote by $\inner{x}{y}{}$ the usual inner product. Moreover, $x\le y$ denotes elementwise inequality. We define $[x]_+=\max\{0,x\}$ and  $[x]_-=-\min\{0,x\}$. The set of probability measures on a set $X$ is denoted by $\mathcal{P}(X)$.

\section{Preliminaries}\label{sec:ProbStatement}

Consider a finite MDP described by a tuple $\mcf{M}_c \triangleq \big( \sspace,\aspace,P,\gamma,\nu_0,c \big)$, where $\sspace=\{x_1,\ldots,x_{\abs{\sspace}}\}$ is the state space, $\aspace=\{a_1,\ldots,a_{\abs{\aspace}}\}$ is the action space, $P:\sspace\times\aspace\mapsto \mathcal{P}(\sspace)$ is the transition law, $\gamma\in(0,1)$ is the discount factor, $\nu_0\in\mcf{P}(\sspace)$ is the initial probability distribution of the system state, and $c:\sspace\times\aspace\mapsto\mbf{R}$ is the one-stage cost function.

The model $\mcf{M}_c$ represents an infinite horizon controlled discrete-time stochastic system whose evolution is described as follows.
At time step $t$, if the system is in state $x_t=x\in\sspace$, and  the action $a_t=a\in\aspace$ is taken, then (i) the cost $c(x,a)$ is incurred, and (ii) the system moves to the next state $x_{t+1}$, which is an $\sspace$-valued random variable with probability distribution $P(\cdot|x,a)$. Once transition into the new state has occurred, a new action is chosen and the process is repeated.

A stationary Markov policy is a map $\pi:\sspace\mapsto\mcf{P}(\aspace)$, and $\pi(a|x)$ denotes the probability of choosing action $a$, while being in state $x$. We denote the space of stationary Markov policies by $\Pi_0$.

Given a policy $\pi\in\Pi_0$, we denote by $\Prob_{\nu_0}^{\pi}$ the induced probability measure on the canonical sample space $\Omega\triangleq(\sspace\times\aspace)^\infty$, i.e.,  $\Prob_{\nu_0}^{\pi}[\cdot]=\textup{Prob}[\cdot\mid\pi,x_0\sim\nu_0]$.
The expectation operator with respect to $\Prob^{\pi}_{\nu_0}$ is denoted by $\Exp^{\pi}_{\nu_0}$.

The optimal control problem is given by
$
	 \min_{\pi\in\Pi_0}\eta_c(\pi),
$
where $\eta_c(\pi) \triangleq \Exp_{\nu_0}^\pi \Big[ \sum_{t=0}^\infty \gamma^t c(x_t,a_t) \Big]$, is the total expected discounted cost of a policy $\pi$.

For every policy $\pi\in\Pi_0$, we define the \emph{$\gamma$-discounted occupancy measure} $\mu ^{\pi}:\sspace\times\aspace\mapsto\mbf{R}_+$, by
$
\mu ^{\pi}(x,a) \triangleq \sum_{t=0}^\infty \gamma^t \Prob_{\nu_0}^{\pi}\left[x_t=x,a_t=a\right].
$
The occupancy measure can be interpreted as the (unnormalized) discounted visitation frequency of state-action pairs when acting according to policy $\pi$. Moreover, it holds that $\eta_c(\pi)=\sum_{x,a}\om (x,a)c(s,a)=\Exp_{\om}[c(x,a)]$.

 \section{Apprenticeship Learning Framework}\label{sec:Method}
 \subsection{Problem statement}
 Consider now the Markov decision model without a cost function, $\mcf{M} \triangleq \big( \sspace,\aspace,P,\gamma,\nu_0 \big)$.
 Assume that instead, we have access to a finite number $m$ of i.i.d sample trajectories $\{(x_0^k,a_0^k,x_1^k,a_1^k,\ldots,x_t^k,a_t^k,\ldots)\}_{k=1}^m$ coming from an expert policy $\pi_E$. Note that the expert policy could also be history-dependent.
 We impose the following assumptions:
 \begin{assumption}[Apprenticeship learning]\label{ass:apprenticeship}~
 	\begin{enumerate}[label=(A\arabic*)]
 		\item\label{ass:A1}
 		$\pi_E$ is a nearly optimal policy for the discounted MDP corresponding to the model $\mcf{M}_{c_{\textup{true}}} = \big( \sspace,\aspace,P,\gamma,\nu_0,c_{\textup{true}}\big)$;
 		
 		\item\label{ass:A2}
 		$c_{\textup{true}}\in\mcf{C}_{\rm lin} = \{\sum_{i=1}^{n_c} w_i \psi_i \mid \norm{w}_\infty \le 1\}$, where $\{\psi_i\}_{i=1}^{n_c}\subset\Re^{\abs{\sspace}\abs{\aspace}}$ are fixed basis vectors, such that $\norm{\psi_i}_\infty \le 1$ for all $i=1,\ldots,n_c$.
 	\end{enumerate}
 \end{assumption}
 The goal of AL is to find a policy $\pi$, such that $\innerprod{\mu ^\pi}{c_{\textup{true}}} \le \innerprod{\mu ^{\pi_E}}{c_{\textup{true}}} $.
 Since the cost function $c_{\textup{true}}$ is unknown, AL algorithms search for a policy $\pi$ that satisfies $\innerprod{\mu ^\pi}{c} \le \innerprod{\mu ^{\pi_E}}{c}$, for all $c\in\mcf{C}_{\rm lin}$. Therefore, an AL algorithm seeks a policy that performs better than the expert across $\mcf{C}_{\rm lin}$, by optimizing the objective 
 \begin{equation}\label{eq:robust_prob}
 	\min_{\pi\in\Pi} \sup_{c\in\mcf{C}_{\rm lin}} \left( \innerprod{\mu ^\pi}{c} - \innerprod{\mu ^{\pi_E}}{c} \right).
 \end{equation}
 We highlight that one can consider other linearly parameterized cost classes, e.g., $\mcf{C}_{\rm lin ,2} = \{\sum_{i=1}^{n_c} w_i \psi_i \mid \norm{w}_2 \le 1\}$~\cite{Abbeel:2004}, or $\mcf{C}_{\rm convex} = \{\sum_{i=1}^{n_c} w_i \psi_i \mid w_i\geq 0,\,\,\sum_{i=1}^{n_c}w_i=1\}$~\cite{Syed:2007,syed2008apprenticeship}. The reasoning and the analysis are similar.
 \subsection{The convex optimization view}
 
 	In the remainder of the paper we will use the following vector notation borrowed from~\cite{Abbasi-Yadkori:2014}.
 	The transition law is a matrix  $P\in\Re_+^{\abs{\sspace}\abs{\aspace}\times\abs{\sspace}}$ so that $\sum_{x'\in\sspace} P_{(x,a),x'}=1$, the initial probability distribution is a vector $\nu_0\in\Re_+^{\abs{\sspace}}$ so that $\norm{\nu_0}_1=1$, and the cost function is a vector $c\in\Re^{\abs{\sspace}\abs{\aspace}}$.
 	Finally, for a stationary Markov policy $\pi\in\Pi_0$ we define the matrix $M^{\pi}\in\Re_+^{\abs{\sspace}\times\abs{\sspace}\abs{\aspace}}$ that encodes $\pi$ as  $M^{\pi}_{x_i,(x_j,a_k)}=\pi(a_k\mid x_i)$, if $i=j$, and $M^{\pi}_{x_i,(x_j,a_k)}=0$ otherwise.

  Next, we will characterize the set of occupancy measures in terms of linear constraint satisfaction.
  To this aim let
  $
   \mcf{F} \triangleq \left\lbrace \mu\in\Re^{\abs{\sspace}\abs{\aspace}} \mid (B-\gamma P)\tpose\mu=\nu_0, \; \mu\ge 0 \right\rbrace,
  $
  where $B\in\{0,1\}^{\abs{\sspace}\abs{\aspace}\times\abs{\sspace}}$ is a binary matrix defined by $B_{(x_i,a_k),x_j}=1$, if $i=j$, and $B_{(x_i,a_k),x_j}=0$ otherwise. The constraints that define the set $\mathcal{F}$ are also known as \emph{Bellman flow constraints}.
  
  \begin{proposition}[{\cite[Theorem 2]{syed2008apprenticeship}}]\label{eq:occup_meas_set}
  	It holds that,
  $
   \mcf{F} = \left\lbrace \mu ^\pi \mid \pi\in\Pi_0 \right\rbrace.
  $
  Indeed, for every $\pi\in\Pi_0$, we have that $\mu ^\pi \in \mcf{F}$.
  Moreover, for every feasible solution $\mu\in \mcf{F}$, we can obtain a stationary Markov policy  $\pi_{\mu}\in\Pi_0$ by
  $
  \pi_{\mu}(a|x) \triangleq \frac{\mu(x,a)}{\sum_{a'\in\aspace}\mu(x,a')}.
  $
  Then, the corresponding induced occupancy measure $\mu ^{\pi_{\mu}}$ satisfies $\mu ^{\pi_\mu}=\mu$.
  \end{proposition}
 
Let $\mbf{\Psi} \triangleq [\psi_1|\ldots|\psi_{n_c}]\in\Re^{\abs{\sspace}\abs{\aspace}\times n_c}$ be the cost basis matrix.
 For a policy $\pi\in\Pi$, we define its \emph{feature expectation vector} as
 \[
 \innerprod{\mu ^\pi}{\mbf{\Psi}} \triangleq \mbf{\Psi}\tpose\mu ^\pi \in\Re^{n_c}.
 \]
 In other words, for every $i=1,\ldots,n_c$,
 \[
 \innerprod{\mu ^\pi}{\mbf{\Psi}}_i = \innerprod{\mu ^\pi}{\psi_i} = \eta_{\psi_i}(\pi).
 \]
 \begin{lemma}\label{lem:sup_cost}
 	For every $\pi\in\Pi$, the following holds
 	\[
 	\sup_{c\in\mcf{C}_{\rm lin}} \left( \innerprod{\mu ^\pi}{c} - \innerprod{\mu ^{\pi_E}}{c} \right) = \norm{\innerprod{\mu ^\pi}{\mbf{\Psi}} - \innerprod{\mu ^{\pi_E}}{\mbf{\Psi}}}_1.
 	\]
 \end{lemma}

 By Lemma~\ref{lem:sup_cost}, we get that \eqref{eq:robust_prob} is equivalent to
 \begin{equation}\label{eq:min_prob}
 	\min_{\pi\in\Pi} \; \norm{ \innerprod{\mu ^\pi}{\mbf{\Psi}} - \innerprod{\mu ^{\pi_E}}{\mbf{\Psi}} }_1.
 \end{equation}
 Note that although the objective function in \eqref{eq:min_prob} is convex in $\mu ^\pi$, the whole program is non-convex in $\pi$.
 However, combining Proposition~\ref{eq:occup_meas_set} with the fact that $\innerprod{\mu ^{\pi}}{c}=\eta_c(\pi)$ for every policy $\pi$, and every cost $c$, we conclude that the AL objective~\eqref{eq:min_prob} can be stated equivalently as a convex optimization program: 
 \begin{equation}\label{eq:min_prob_cvx}
 	\min_{\mu\in \mcf{F}} \norm{ \innerprod{\mu}{\mbf{\Psi}} - \innerprod{\mu ^{\pi_E}}{\mbf{\Psi}} }_1.
 \end{equation}
  Note that the $\abs{\sspace}\abs{\aspace}$ linear constraints given by $\mu\ge 0$ ensure that $\mu$ is a nonnegative measure, while the $\abs{\sspace}$ linear constraints given by $(B-\gamma P)\tpose\mu=\nu_0$ ensure that $\mu$ is an occupancy measure generated by a stationary Markov policy.

 \section{Algorithm and main result}
  In practice we do not have access to the whole policy $\pi_E$, but instead can observe \iid trajectory samples distributed according to $\Prob_{\nu_0} ^{\pi_E}$.
  For a multi-sample $\{(x_0^k,a_0^k,x_1^k,a_1^k,\ldots,x_t^k,a_t^k,\ldots)\}_{k=1}^m \sim (\Prob_{\nu_0}^{\pi_E})^m$ consider the Monte Carlo approximation $\widehat{\innerprod{\mu ^{\pi_E}}{\mbf{\Psi}}}\in\Re^{n_c}$ of the expert feature expectation vector, i.e., for each $i=1,\ldots,n_c$,
  \[
  \widehat{\innerprod{\mu ^{\pi_E}}{\mbf{\Psi}}}_i = \widehat{\innerprod{\mu ^{\pi_E}}{\psi_i}} \triangleq \frac{1}{m} \sum_{t=0}^\infty \sum_{j=1}^m \gamma^t \psi_i(x_t^j,a_t^j).
  \]
 Moreover, under Assumption~\ref{ass:A2}, the following is a pointwise bound on $\Omega^m$:
  \begin{equation}\label{eq:bound_norm}
  \norm{ \widehat{\innerprod{\mu ^{\pi_E}}{\mbf{\Psi}}} }_\infty \le 1/(1-\gamma).
  \end{equation}
   We are interested in optimizing the empirical  convex objective for large-scale MDPs:
  \begin{equation}\label{eq:final_prob}
  \min_{\mu\in \mcf{F}} \norm{ \innerprod{\mu}{\mbf{\Psi}} - \widehat{\innerprod{\mu ^{\pi_E}}{\mbf{\Psi}}} }_1,
  \end{equation}
  which is a random convex program on $(\Omega^m,(\Prob_{\nu_0}^{\pi_E})^m)$.

 Our main aim is (i) to provide a computationally efficient algorithm whose complexity does not grow with the size of the state and action spaces, and (ii) to obtain explicit probabilistic performance bounds on the quality of the extracted solution.
 To this end, we will design and analyze the AL analogue of the algorithm proposed in \cite{Abbasi-Yadkori:2014} for the forward average-cost MDP problem. Most of the tools from the forward MDP setting~ \cite{Abbasi-Yadkori:2014,Yadkori2019} can be used for the AL formulation with the appropriate modifications. We will present the main reasoning and results in this section with proofs presented in the appendix.

 As the first step, instead of optimizing over the whole space $\Re^{\abs{\sspace}\abs{\aspace}}$, we optimize over the linear hull of a small number of selected feature vectors $\{\phi_i\}_{i=1}^d \subset \Re^{\abs{\sspace}\abs{\aspace}}$.
 In this way, we reduce significantly the number of optimization variables.
 Let $\mbf{\Phi} \triangleq [\phi_1|\ldots|\phi_d]\in\Re^{\abs{\sspace}\abs{\aspace}\times d}$ be the feature matrix.
 The corresponding reduced program is
 \begin{equation}\label{eq:reduced_prob}
 	\min_{\theta\in\Re^d: \mbf{\Phi}\theta\in \mcf{F}} \norm{ \mbf{\Psi}\tpose\mbf{\Phi}\theta - \widehat{\innerprod{\mu ^{\pi_E}}{\mbf{\Psi}}} }_1.
 \end{equation}
 
 Note that for an arbitrary vector $u\in\Re^{\abs{\sspace}\abs{\aspace}}$, which is not necessarily in $ \mcf{F}$, we can still define a policy $\pi_u\in\Re^{\abs{\sspace}\abs{\aspace}}$ by
 $
 \pi_u(a|x) = \frac{[u(x,a)]_+}{\sum_{a'\in\aspace}[u(x,a')]_+}.
 $
 If $u(x,a)\le 0$, for all $a\in\aspace$, we let $\pi_u(\cdot\mid x)$ be the uniform distribution~\cite{Yadkori-AISTATS}. Then, one has that $\mu ^{\pi_u}=u$ if and only if $u\in \mcf{F}$.
 In the general case, the following lemma, which is the discounted cost analogue of \cite[\Lem~2]{Abbasi-Yadkori:2014}, quantifies how close the generated occupancy measure $\mu ^{\pi_u}$ is to $u$ according to the degree of constraint violation.

 \begin{lemma}\label{lem:occup_meas_dist}
 	For any $u\in\Re^{\abs{\sspace}\abs{\aspace}}$, it holds that
 	$
 	\norm{\mu ^{\pi_u}-u}_1 \le \frac{2\norm{[u]_-}_1 + \norm{(B-\gamma P)\tpose u-\nu_0}_1}{1-\gamma}.
 	$
 \end{lemma}
 
 For any $\theta\in\Re^d$, we define $\pi_\theta \triangleq \pi_{\mbf{\Phi}\theta}$ and $\mu_\theta \triangleq \mu ^{\pi_{\mbf{\Phi}\theta}}$.
 As already discussed, $\mu_\theta = \mbf{\Phi}\theta$ if and only if $\theta$ is feasible for \eqref{eq:reduced_prob}.
 In the general case, one can bound the distance between the occupancy measure $\mu_\theta$ and the vector $\mbf{\Phi}\theta$ by applying Lemma~\ref{lem:occup_meas_dist}.
 
 Let $(\rho,\lambda)>0$ be positive constants, $\Theta \triangleq \{ \theta\in\Re^d \mid \norm{\theta}_2 \le \rho \}$, and $\Pi_\Theta:\Re^d\mapsto\Theta$ the Euclidean projection onto $\Theta$.
 Consider the following surrogate loss function which is obtained by adding a positive multiple of the constraint violations to the initial objective function:
 
 \begin{align*}
 	\mcf{L}(\theta) &\triangleq \norm{ \mbf{\Psi}\tpose\mbf{\Phi}\theta - \widehat{\innerprod{\mu ^{\pi_E}}{\mbf{\Psi}}} }_1 + \lambda\underbrace{\norm{[\mbf{\Phi}\theta]_-}_1}_{\coloneqq V_1(\theta)} + \lambda\underbrace{\norm{(B-\gamma P)\tpose (\mbf{\Phi}\theta)-\nu_0}_1}_{\coloneqq V_2(\theta)} \\
 	&= \sum_{i=1}^{n_c} \abs{\mbf{\Psi}_{:,i}\tpose\mbf{\Phi}\theta - \widehat{\innerprod{\mu ^{\pi_E}}{\psi_i}}} + \lambda\sum_{(x,a)\in\sspace\times\aspace}[\mbf{\Phi}_{(x,a),:}\theta]_- + \lambda \sum_{x\in\sspace} \abs{(B-\gamma P)_{:,x}\tpose \mbf{\Phi}\theta-\nu_0(x)}.
 \end{align*}

 We are interested in the reduced unconstrained convex optimization program of the form
 $
 \min_{\theta\in\Theta} \mcf{L}(\theta).
 $
 
 A subgradient of $\mcf{L}$ at $\theta$ is given by
 
 \begin{align*}
 	\nabla_\theta \mcf{L}(\theta) &= \sum_{i=1}^{n_c} (\mbf{\Phi}\tpose\mbf{\Psi}_{:,i})\,\sign \left( \mbf{\Psi}_{:,i}\tpose\mbf{\Phi}\theta - \widehat{\innerprod{\mu ^{\pi_E}}{\psi_i}} \right) \\
 	&\phantom{{}=} + \lambda \sum_{x\in\sspace}(\mbf{\Phi}\tpose(B-\gamma P)_{:,x}) \, \sign \left( (B-\gamma P)_{:,x}\tpose \mbf{\Phi}\theta-\nu_0(x) \right) \\
 	&\phantom{{}=} - \lambda\sum_{(x,a)\in\sspace\times\aspace}(\mbf{\Phi}_{(x,a),:}\tpose) \ones_{\{\mbf{\Phi}_{(x,a),:}\theta<0\}}.
 \end{align*}

 Suppose that $q_1\in\mcf{P}(\sspace\times\aspace)$ and $q_2\in\mcf{P}(\sspace)$ assign to each element a strictly positive probability.
 We propose a method for AL shown in Algorithm~\ref{alg:sgd-al}.
 It uses an unbiased estimate of $\nabla_\theta \mcf{L}(\theta)$ for fixed expert trajectory samples, \ie
 
 \begin{align}\label{eq:subgrad}
 	\begin{split}
 		g_t(\theta) &= \sum_{i=1}^{n_c} (\mbf{\Phi}\tpose \mbf{\Psi}_{:,i}) \, \sign \left(\mbf{\Psi}_{:,i}\tpose\mbf{\Phi}\theta - \widehat{\innerprod{\mu ^{\pi_E}}{\psi_i}} \right) \\
 		&\phantom{{}=} + \lambda \frac{\mbf{\Phi}\tpose (B-\gamma P)_{:,y^{(t)}}}{q_2(y^{(t)})} \, \sign \left( (B-\gamma P)_{:,y^{(t)}}\tpose\mbf{\Phi}\theta-\nu_0(y^{(t)}) \right) \\
 		&\phantom{{}=} - \lambda \frac{\mbf{\Phi}_{(x^{(t)},a^{(t)}),:}\tpose}{q_1(x^{(t)},a^{(t)})}\ones_{\{\mbf{\Phi}_{(x^{(t)},a^{(t)}),:}\theta<0\}},
 	\end{split}
 \end{align}

 where $(x^{(t)},a^{(t)})\sim q_1$ and $y^{(t)}\sim q_2$.
 
 \begin{algorithm}[t]
 	\caption{Stochastic subgradient descent for apprenticeship learning (SGD-AL).}
 	\label{alg:sgd-al}
 	\begin{algorithmic}[1]
 		\State \textbf{given} cost matrix $\mbf{\Psi}$, feature matrix $\mbf{\Phi}$, number of expert samples $m$, number of iterations $T$, learning rate $\eta>0$, radius $\rho>0$, regularization parameter $\lambda>0$
 		\State Set $\theta_0=0$
 		\State Sample $\{(x_0^k,a_0^k,x_1^k,a_1^k,\ldots,x_t^k,a_t^k,\ldots)\}_{k=1}^m\sim(\Prob_{\nu_0}^{\pi_E})^m$
 		\For{$t=1,\ldots,T$}
 		\State Sample $(x^{(t)},a^{(t)})\sim q_1$ and $y^{(t)}\sim q_2$
 		\State Compute $g_t(\theta_t)$ via \eqref{eq:subgrad}
 		\State Update $\theta_{t+1} = \Pi_\Theta \big( \theta_t - \eta g_t(\theta_t) \big)$
 		\EndFor
 		\State $\hat{\theta}_T=\frac{1}{T} \sum_{t=1}^T \theta_t$
 		\State Return $\pi_{\hat{\theta}_T}$
 	\end{algorithmic}
 \end{algorithm}
 
  Regret bounds for Algorithm~\ref{alg:sgd-al} can be obtained by using results from the stochastic convex optimization literature and statistical learning theory.
 \begin{assumption}\label{ass:feature_mat}
 	All entries of the feature matrix $\mbf{\mbf{\Phi}}$ are positive, \ie every feature vector $\phi_i$ is a measure which assigns a non-zero value to each pair $(x,a)$. Moreover, $\norma{\mbf{\Phi}}{1}=\frac{1}{1-\gamma}$.
 \end{assumption}
 We define the following constants:
 \begin{align*}
 	C_1 &\triangleq \max_{(x,a)\in\sspace\times\aspace} \frac{\norm{\mbf{\Phi}_{(x,a),:}\tpose}_2}{q_1(x,a)}, \\
 	C_2 &\triangleq \max_{x\in\sspace} \frac{\norm{\mbf{\Phi}\tpose (B-\gamma P)_{:,x}}_2}{q_2(x)}.
 \end{align*}
 These constants appear in our performance bounds.
 We would like to choose appropriate distributions so that $C_1$ and $C_2$ are small, since they  appear in the error bound. We refer the  reader to~\cite{Abbasi-Yadkori:2014} for a thorough discussion on the choice of the distributions.
 
 
 Observe from \eqref{eq:subgrad} that for all $\theta\in\Theta$ we have the following bound:
 \begin{equation}\label{eq:bound_subgrad}
 	\norm{g_t(\theta)}_2 \le \norm{\mbf{\Phi}}_2 \sum_{i=1}^{n_c} \norm{\psi_i}_2 + \lambda(C_1+C_2) \eqqcolon K.
 \end{equation}
 
 
 
 \begin{theorem}\label{thm:regret}
 	Let $\eps\in(0,1)$, $\delta\in(0,1)$, $\rho>0$, $\lambda=1/\eps$, $m\ge\frac{32 n_c^2\log(\frac{4n_c}{\delta})}{(1-\gamma)\eps^2}$, $T\ge \frac{4\rho^2}{\eps^2}\left( \frac{2\norm{\mbf{\Psi}}_\infty}{\lambda (1-\gamma)} + 1 \right)^2 \Delta^2$ with
 	$
 	\Delta \triangleq K + \sqrt{10\log\frac{2}{\delta}} + \sqrt{5d\log (1+\frac{\rho^2 T}{d})},
 	$
 	and $\eta = \rho/(K\sqrt{T})$.
 	Then, with probability at least $1-\delta$, Algorithm~\ref{alg:sgd-al} generates $\pi_{\hat{\theta}_T}$ so that for all $\theta\in\Theta$,
 	
 	\begin{align*}
 		\norm{ \mbf{\Psi}\tpose\mu_{\hat{\theta}_T} - \innerprod{\mu ^{\pi_E}}{\mbf{\Psi}} }_1 &\le \norm{ \mbf{\Psi}\tpose\mu_\theta - \innerprod{\mu ^{\pi_E}}{\mbf{\Psi}} }_1 + \left( \frac{4\norm{\mbf{\Psi}}_\infty}{1-\gamma} + \frac{1}{\eps} \right) ( V_1(\theta)+V_2(\theta) ) \\
 		&\phantom{{}=} + \left( \frac{2\norm{\mbf{\Psi}}_\infty}{1-\gamma} \right) \left(\norm{\mbf{\Psi}}_\infty \norm{\mbf{\Phi}}_1\rho\sqrt{d} + \frac{n_c}{1-\gamma} \right) \eps + \eps.
 	\end{align*}

 \end{theorem}


 \bibliographystyle{plain}
\bibliography{refs}
 \newpage
 \appendix
  \section{Proofs}
  
    \begin{proofof}{Proof of Lemma 1}
    	Fix a $\pi\in\Pi$.
    	Then, for every $c=\sum_{i=1}^{n_c} w_i \psi_i$ with $\norm{w}_\infty \le 1$, it holds that
    	\begin{align*}
    	\innerprod{\mu ^\pi}{c} - \innerprod{\mu ^{\pi_E}}{c} &= \sum_{i=1}^{n_c} w_i \left( \innerprod{\mu ^\pi}{\psi_i} - \innerprod{\mu ^{\pi_E}}{\psi_i} \right) \\
    	&\le \norm{w}_\infty \norm{ \innerprod{\mu ^\pi}{\mbf{\Psi}} - \innerprod{\mu ^{\pi_E}}{\mbf{\Psi}} }_1 \\
    	&\le \norm{ \innerprod{\mu ^\pi}{\mbf{\Psi}} - \innerprod{\mu ^{\pi_E}}{\mbf{\Psi}} }_1,
    	\end{align*}
    	and thus,
    	\[
    	\sup_{c\in\mcf{C}_{\rm lin}} \left( \innerprod{\mu ^\pi}{c} - \innerprod{\mu ^{\pi_E}}{c} \right)\le \norm{\innerprod{\mu ^\pi}{\mbf{\Psi}} - \innerprod{\mu ^{\pi_E}}{\mbf{\Psi}} }_1.
    	\]
    	Next, let $\tilde{w}\in\Re^{n_c}$ be defined by
    	\[
    	\tilde{w}_i = \textup{sign} \left(\innerprod{\mu ^\pi}{\psi_i} - \innerprod{\mu ^{\pi_E}}{\psi_i} \right).
    	\]
    	Then for $\tilde{c} \triangleq \sum_{i=1}^{n_c} \tilde{w}_i \psi_i$, we have $\tilde{c}\in\mcf{C}_{\rm lin}$ and
    	\[
    	\innerprod{\mu ^\pi}{\tilde{c}} - \innerprod{\mu ^{\pi_E}}{\tilde{c}} = \norm{ \innerprod{\mu ^\pi}{\mbf{\Psi}} - \innerprod{\mu ^{\pi_E}}{\mbf{\Psi}} }_1,
    	\]
    	which proves that
    	\[
    	\sup_{c\in\mcf{C}_{\rm lin}} \left(\innerprod{\mu ^\pi}{c}{}-\innerprod{\mu ^{\pi_E}}{c} \right) \ge \norm{ \innerprod{\mu ^\pi}{\mbf{\Psi}} - \innerprod{\mu ^{\pi_E}}{\mbf{\Psi}} }_1.
    	\]
    	This concludes the proof.
    \end{proofof}

 In case of a stationary Markov policy, the induced discounted occupancy measure has the following form.
 \begin{lemma}\label{lem:occup_meas}
 	Let $\pi\in\Pi_0$ be a stationary Markov policy.
 	Then for all $x\in\sspace$, $a\in\aspace$ and $t\in\Nat_0$, it holds that
 	$
 	\Prob_{\nu_0}^\pi[x_t=x,a_t=a] = \left[ \nu_0\tpose M^{\pi}(P M^\pi)^t \right]_{(x,a)}.
 	$
 	In particular, $(\mu ^{\pi})\tpose = \sum_{t=0}^\infty \gamma^t\nu_0\tpose M^{\pi}(PM^{\pi})^t.$
 \end{lemma}
 \begin{proofof}{Proof}
 	For $t=0$ we have
 	
 	\begin{align*}
 	\Prob_{\nu_0}^{\pi}[x_0=x, a_0=a] &= \nu_0(x)\pi(a|x)=[\nu_0^\top M^{\pi}]_{(x,a)}.
 	\end{align*}
 	
 	Next, assume that the result holds for $t-1$.
 	Then,
 	
 	\begin{align*}
 	\Prob_{\nu_0}^{\pi}[x_t=x,a_t=a] &= \Prob_{\nu_0}^{\pi}[a_t=a \mid x_t=x]\Prob_{\nu_0}^{\pi}[x_t=x] \\
 	&= \pi(a|x) \sum_{x'\in\sspace}\sum_{a'\in\aspace} P_{(x',a'),x}\Prob_{\nu_0}^{\pi}[x_{t-1}=x',a_{t-1}=a'].
 	\end{align*}
 	
 	By the induction assumption, we conclude that
 	
 	\begin{align*}
 	\Prob_{\nu_0}^{\pi}[x_t=x,a_t=a] &= \pi(a|x) \sum_{x'\in\sspace}\sum_{a'\in\aspace}\left[\nu_0^\top M^{\pi}(PM^{\pi})^{t-1}\right]_{(x',a')} P_{(x',a'),x} \\
 	&= \pi(a|x)\left[\nu_0^\top M^{\pi}(PM^{\pi})^{t-1}P\right]_{x} \\
 	&= \left[\nu_0^\top M^{\pi}(PM^{\pi})^t\right]_{(x,a)}.
 	\qedhere
 	\end{align*}
 	
 \end{proofof}
 
  \begin{proofof}{Proof of Lemma 2}
  
  We provide a refined proof and bound similar to~\cite[Lemma 13]{Yadkori2019}.
  	We have
  	
  	\begin{align*}
  	\norm{\underbrace{(B-\gamma P)\tpose [u]_+-\nu_0}_{=: -w}}_1 &\le \norm{(B-\gamma P)\tpose [u]_-}_1+\norm{(B-\gamma P)\tpose u-\nu_0}_1 \\
  	&\le (1+\gamma)\norm{[u]_-}_1 + \norm{(B-\gamma P)\tpose u-\nu_0}_1,
  	\end{align*}

  	where we have used the fact that $\norm{B\tpose}_1 = \norm{B}_\infty = 1$ and $\norm{P\tpose}_1 = \norm{P}_\infty = 1$.
  	By virtue of Lemma~\ref{lem:occup_meas}, we have
  	
  	\begin{align*}
  	(\mu ^{\pi_u})\tpose &= \sum_{t=0}^\infty \gamma^t \nu_0\tpose M^{\pi_u}(PM^{\pi_u})^t \\
  	&= \sum_{t=0}^\infty \gamma^t (w +(B-\gamma P)\tpose [u]_+)\tpose M^{\pi_u}(PM^{\pi_u})^t \\
  	&= \sum_{t=0}^\infty \gamma^t w\tpose M^{\pi_u}(P M^{\pi_u})^t + \sum_{t=0}^\infty \gamma^t [u]_+\tpose (PM^{\pi_u})^t - \sum_{t=0}^\infty \gamma^{t+1} [u]_+\tpose (PM^{\pi_u})^{t+1} \\
  	&= \sum_{t=0}^\infty \gamma^t w\tpose M^{\pi_u}(P M^{\pi_u})^t + [u]_+\tpose,
  	\end{align*}

  	where in the third equality we used $[u]_+\tpose B M^{\pi_u} = [u]_+\tpose$.
  	Therefore,
  	
  	\begin{align*}
  	\norm{\mu ^{\pi_u}-[u]_+}_1 &= \norm{\sum_{t=0}^\infty \gamma^t((M^{\pi_u})\tpose P\tpose)^t (M^{\pi_u})\tpose w}_1 \\
  	&\le \sum_{t=0}^\infty \gamma^t \norm{M^{\pi_u}}_\infty^t \norm{P}_\infty^t \norm{M^{\pi_u}}_\infty \norm{w}_1 \\
  	&\le \frac{1}{1-\gamma} \left( (1+\gamma)\norm{[u]_-}_1 + \norm{(B-\gamma P)\tpose u-\nu_0}_1 \right),
  	\end{align*}

  	where in the last inequality we used $\norm{M^{\pi_u}}_\infty = \norm{B}_\infty = \norm{P}_\infty = 1$.
  	Finally, the triangle inequality gives
  	
  	\[
  	\norm{\mu ^{\pi_u}-u}_1 \le \norm{\mu ^{\pi_u} - [u]_+}_1 + \norm{[u]_-}_1 \le \frac{2\norm{[u]_-}_1 + \norm{(B-\gamma P)\tpose u-\nu_0}_1}{1-\gamma}.
  	\qedhere
  	\]
  	
  \end{proofof}

    \begin{proofof}{Proof of Theorem 1}
    	The proof combines techniques presented in the proofs of~\cite[Theorem 2]{Abbasi-Yadkori:2014} and~\cite[Lemma 14]{Yadkori2019} and the Hoeffding's bound.
    	 
    	We first fix an expert trajectory multi-sample $\{(x_0^k,a_0^k,x_1^k,a_1^k,\ldots,x_t^k,a_t^k,\ldots)\}_{k=1}^m\sim(\Prob_{\nu_0}^{\pi_E})^m$.
    	Then, by virtue of \cite[\Thm~3]{Abbasi-Yadkori:2014} and by the uniform bound of the unbiased subgradient estimates \eqref{eq:bound_subgrad}, we get that if the learning rate is $\eta = \rho/(K\sqrt{T})$, then with probability at least $1-\delta/2$ (the corresponding probability space is $((\sspace\times\aspace)\tpose\times\sspace\tpose),q_1\tpose\otimes q_2\tpose$),
    	
    	\begin{equation}\label{eq:suboptimality}
    	\mcf{L}(\hat{\theta}_T) - \min_{\theta\in\Theta}\mcf{L}(\theta) \le \frac{\rho K}{\sqrt{T}} + \sqrt{ \frac{1+4\rho^2T}{T^2} \left( 2\log\frac{2}{\delta}+d\log\left(1 + \frac{\rho^2 T}{d} \right) \right) }.
    	\end{equation}

    	Integrating over the whole probability space $(\Omega^m,(\Prob_{\nu_0}^{\pi_E})^m)$, we conclude that \eqref{eq:suboptimality} holds with probability at least $1-\delta/2$, where the corresponding probability space is $(\Omega^m\times(\sspace\times\aspace)\tpose\times\sspace\tpose),(\Prob_{\nu_0}^{\pi_E})^m\otimes q_1\tpose\otimes q_2\tpose$).

    	Substituting $\mcf{L}(\hat{\theta}_T)$ and $\mcf{L}(\theta)$ by their definitions, and using the inequality $\sqrt{a+b} \le \sqrt{a} + \sqrt{b}$, we obtain that with probability at least $1-\delta/2$, for all $\theta\in \Theta$,
    		\begin{equation}\label{A1}
    	\norm{\mbf{\Psi}\tpose\mbf{\Phi}\hat{\theta}_T - \widehat{\innerprod{\mu ^{\pi_E}}{\mbf{\Psi}}}}_1 + \lambda V_1(\hat{\theta}_T)+\lambda V_2(\hat{\theta}_T) \le \norm{ \mbf{\Psi}\tpose\mbf{\Phi}\theta - \widehat{\innerprod{\mu ^{\pi_E}}{\mbf{\Psi}}} }_1 + \lambda V_1(\theta) + \lambda V_2(\theta) + \frac{\rho}{\sqrt{T}}\Delta.
    	\end{equation}
    	For all multi-samples that~(\ref{A1}) holds, and for all $\theta\in\Theta$,
    	\begin{align}
    	\norm{\mbf{\Psi}\tpose \mu_{\hat{\theta}_T} - \widehat{\innerprod{\mu ^{\pi_E}}{\mbf{\Psi}}}}_1 &\le \norm{\mbf{\Psi}\tpose \mu_{\hat{\theta}_T} - \mbf{\Psi}\tpose\mbf{\Phi} \hat{\theta}_T}_1 + \norm{\mbf{\Psi}\tpose\mbf{\Phi} \hat{\theta}_T - \widehat{\innerprod{\mu ^{\pi_E}}{\mbf{\Psi}}}}_1 \nonumber \\
    	&\le \norm{\mbf{\Psi}}_\infty \frac{2 V_1(\hat{\theta}_T)+V_2(\hat{\theta}_T)}{1-\gamma} + \norm{\mbf{\Psi}\tpose\mbf{\Phi}\theta - \widehat{\innerprod{\mu ^{\pi_E}}{\mbf{\Psi}}}}_1 \nonumber \\
    	&\phantom{{}\le} + \lambda V_1(\theta) + \lambda V_2(\theta) + \frac{\rho}{\sqrt{T}}\Delta, \label{A2}
    	\end{align}
    	where we used the triangle inequality in the first step, and Lemma~\ref{lem:occup_meas_dist} together with the bound~(\ref{A1}) in the second.
    	Moreover, by~(\ref{A1}),  
    	\begin{equation}\label{A3}
    	V_1(\hat{\theta}_T) + V_2(\hat{\theta}_T) \le \frac{1}{\lambda} \left( \norm{\mbf{\Psi}}_\infty \norm{\mbf{\Phi}}_1 \rho\sqrt{d} + \frac{n_c}{1-\gamma} \right) +  V_1(\theta) + V_2(\theta)  + \frac{\rho}{\lambda\sqrt{T}}\Delta,
    	\end{equation}
    	where we used that $\norma{\theta}{1}\le\rho\sqrt{d}$, and the pointwise bound $ \norm{ \widehat{\innerprod{\mu ^{\pi_E}}{\mbf{\Psi}}} }_1 \le n_c\,\norm{ \widehat{\innerprod{\mu ^{\pi_E}}{\mbf{\Psi}}} }_\infty  \le n_c/(1-\gamma)$.
    	
    	
    	Once more, by the triangle inequality and Lemma~\ref{lem:occup_meas_dist}, we get
    	
    	\begin{align}
    	\norm{\mbf{\Psi}\tpose\mbf{\Phi}\theta - \widehat{\innerprod{\mu ^{\pi_E}}{\mbf{\Psi}}}}_1 &\le \norm{\mbf{\Psi}\tpose\mbf{\Phi}\theta - \mbf{\Psi}\tpose \mu_\theta}_1 + \norm{\mbf{\Psi}\tpose\mu_\theta - \widehat{\innerprod{\mu ^{\pi_E}}{\mbf{\Psi}}}}_1 \nonumber\\
    	&\le \norm{\mbf{\Psi}}_\infty \frac{2 V_1(\theta) + V_2(\theta)}{1-\gamma} + \norm{\mbf{\Psi}\tpose\mu_\theta - \widehat{\innerprod{\mu ^{\pi_E}}{\mbf{\Psi}}}}_1. \label{A4}
    	\end{align}

    	Therefore, by combining~\eqref{A2},\eqref{A3} and \eqref{A4}, we get that with probability at least $1-\delta/2$, for all $\theta\in\Theta$,
    	
    	\begin{align*}
    	\norm{\mbf{\Psi}\tpose\mu_{\hat{\theta}_T} - \widehat{\innerprod{\mu ^{\pi_E}}{\mbf{\Psi}}}}_1 &\le \norm{\mbf{\Psi}\tpose\mu_\theta - \widehat{\innerprod{\mu ^{\pi_E}}{\mbf{\Psi}}}}_1 + \frac{2\norm{\mbf{\Psi}}_\infty}{\lambda(1-\gamma)} \left( \norm{\mbf{\Psi}}_\infty \norm{\mbf{\Phi}}_1 \rho\sqrt{d} + \frac{n_c}{1-\gamma} \right) \\
    	&\phantom{{}\le} + \left( \frac{4\norm{\mbf{\Psi}}_\infty}{1-\gamma} + \lambda \right) (V_1(\theta) + V_2(\theta)) + \left( \frac{2\norm{\mbf{\Psi}}_\infty}{\lambda (1-\gamma)} + 1 \right) \frac{\rho}{\sqrt{T}} \Delta.
    	\end{align*}
    	
    	For $T\ge\frac{4\rho^2}{\eps^2} \left( \frac{2\norm{\mbf{\Psi}}_\infty}{\lambda(1-\gamma)} + 1 \right)^2 \Delta^2$ and $\lambda=1/\eps$, it follows that with probability at least $1-\delta/2$, for all $\theta \in\Theta$,
    	
    	\begin{align*}
    	\norm{\mbf{\Psi}\tpose\mu_{\hat{\theta}_T} - \widehat{\innerprod{\mu ^{\pi_E}}{\mbf{\Psi}}}}_1 &\le \norm{\mbf{\Psi}\tpose\mu_\theta - \widehat{\innerprod{\mu ^{\pi_E}}{\mbf{\Psi}}}}_1 + \left( \frac{4\norm{\mbf{\Psi}}_\infty}{1-\gamma} + \frac{1}{\eps} \right) (V_1(\theta) + V_2(\theta)) \\
    	&\phantom{{}\le} + \left( \frac{2\norm{\mbf{\Psi}}_\infty}{1-\gamma} \right) \left(\norm{\mbf{\Psi}}_\infty \norm{\mbf{\Phi}}_1 \rho\sqrt{d} + \frac{n_c}{1-\gamma} \right) \eps + \eps/2.
    	\end{align*}

    	We conclude the proof by using Hoeffding's inequality with confidence $\delta/(2n_c)$ and approximation accuracy $\eps/(4n_c)$.
    	In particular, we have that for $m\ge\frac{32 n_c\log(4n_c/\delta)}{2(1-\gamma)\eps^2}$ and for all $i=1,\ldots,n_c$,
    	\[
    	\left| \innerprod{\mu ^{\pi_E}}{\psi_i} - \widehat{\innerprod{\mu ^{\pi_E}}{\psi_i}} \right| \le \eps/(4n_c),
    	\]
    	with probability $(\Prob_{\nu_0}^{\pi_E})^m$ at least $(1-\delta/(2n_c))$.
    	Note that under Assumption~\ref{ass:A2}, it holds that $\sum_{t=0}^\infty \gamma^t\psi_i(x_t,a_t) \le 1/(1-\gamma)$ for all $(x_t,a_t)\in\sspace\times\aspace$ and for all $i=1,\ldots,n_c$.
    	
    	
    	A union bound gives that for $m\ge\frac{32 n_c\log(\frac{4n_c}{\delta})}{2(1-\gamma)\eps^2}$,
    	\[
    	\norm{\innerprod{\mu ^{\pi_E}}{\mbf{\Psi}} - \widehat{\innerprod{\mu ^{\pi_E}}{\mbf{\Psi}}}}_1 \le \eps/4,
    	\]
    	with probability $(\Prob_{\nu_0}^{\pi_E})^m$ at least $(1-\delta/2)$.
    	Integrating over the whole space $\left( (\sspace\times\aspace)\tpose\times\sspace\tpose,q_1\tpose\otimes q_2\tpose \right)$ we have the same statement with probability $(\Prob_{\nu_0}^{\pi_E})^m\otimes q_1\tpose\otimes q_2\tpose$ at least $(1-\delta/2)$.
    	
    	Finally, a simple union bound concludes the proof. $\hat{\overline{\delta}}$
    \end{proofof}

\end{document}